\documentclass{article}

\PassOptionsToPackage{numbers}{natbib}

\usepackage[preprint]{neurips_2023}

\usepackage[utf8]{inputenc} 
\usepackage[T1]{fontenc}    
\usepackage{hyperref}       
\usepackage{url}            
\usepackage{booktabs}       
\usepackage{amsfonts}       
\usepackage{nicefrac}       
\usepackage{microtype}      
\usepackage{xcolor}         
\usepackage{multirow}
\usepackage{graphicx}
\usepackage{pifont}
\usepackage{color}
\usepackage[normalem]{ulem}
\usepackage{tablefootnote}
\usepackage[ruled]{algorithm2e}
\usepackage{boxedminipage}
\usepackage{amsmath}
\usepackage{makecell}
\usepackage{soul}
\usepackage{wrapfig} 

\definecolor{my_green}{RGB}{51,102,0}
\definecolor{my_yellow}{RGB}{255,165,0}
\definecolor{my_red}{RGB}{204, 0, 0}
\definecolor{error_red}{RGB}{241,156,153}
\definecolor{pass_green}{RGB}{185,224,165}

\newcommand{\cmark}{\textcolor{my_green}{\ding{51}}} 
\newcommand{\xmark}{\textcolor{my_red}{\ding{55}}} 

\newcommand{\ttab}{\textcolor{white}{xxxx}}

\definecolor{shadecolor}{RGB}{237,237,237}
\newcommand{\mybox}[1]{\vspace{0.1mm}\par\noindent\colorbox{shadecolor}
{\parbox{\dimexpr\textwidth/2-2\fboxsep\relax}{\vspace{-0.2mm}#1\vspace{-0.2mm}}}\vspace{0.1mm}}
\newcommand{\myboxx}[1]{\vspace{0.1mm}\par\noindent\colorbox{shadecolor}
{\parbox{\dimexpr\textwidth-2\fboxsep\relax}{\vspace{-0.2mm}#1\vspace{-0.2mm}}}\vspace{0.1mm}}

\newcommand{\ourmethod}{RestGPT}

\title{RestGPT: Connecting Large Language Models \\with Real-World RESTful APIs}

\author{%
  Yifan Song$^1$,
  Weimin Xiong$^1$,
  Dawei Zhu$^1$,
  Wenhao Wu$^1$,
  Han Qian$^2$,
  Mingbo Song$^2$\\
  \textbf{Hailiang Huang}$^2$,
  \textbf{Cheng Li}$^3$,
  \textbf{Ke Wang}$^3$,
  \textbf{Rong Yao}$^3$,
  \textbf{Ye Tian}$^3$,
  \textbf{Sujian Li}$^1$\footnotemark[1]\\
  $^1$School of Computer Science, Peking University\quad\\
  $^2$School of Electronics Engineering and Computer Science, Peking University\quad\\
  $^3$Huawei Technologies\quad\\
  \texttt{\{yfsong, lisujian\}@pku.edu.cn} \\
  \vspace{-3mm}\\
  \textbf{\texttt{\href{https://restgpt.github.io}{\textcolor{magenta}{https://restgpt.github.io}}}}
}

\begin{document}

\renewcommand{\thefootnote}{\fnsymbol{footnote}}
\footnotetext[1]{Corresponding author.}
\renewcommand{\thefootnote}{\arabic{footnote}}

\maketitle

\begin{abstract}
Tool-augmented large language models (LLMs) have achieved remarkable progress in tackling a broad range of tasks.
However, existing methods are mainly restricted to specifically designed tools and fail to fulfill complex instructions, having great limitations when confronted with real-world scenarios.
In this paper, we explore a more realistic scenario by connecting LLMs with RESTful APIs, which adhere to the widely adopted REST software architectural style for web service development.
To address the practical challenges of tackling complex instructions, we propose \ourmethod{}, which exploits the power of LLMs and conducts a coarse-to-fine online planning mechanism to enhance the abilities of task decomposition and API selection.
\ourmethod{} also contains an API executor tailored for calling RESTful APIs, which can meticulously formulate parameters and  parse API responses.  
To fully evaluate the performance of \ourmethod{}, we  propose RestBench, a high-quality benchmark which consists of two real-world scenarios and human-annotated instructions with gold solution paths.
Experiments show that \ourmethod{} is able to achieve impressive results in complex tasks and has strong robustness, which paves a new way towards AGI.

\end{abstract}

\section{Introduction}

Large language models (LLMs), such as GPT-3 \citep{brown2020language} and ChatGPT \citep{openai2022chatgpt}, have shown various emergent abilities, including in-context learning \citep{brown2020language,dong2022survey}, reasoning \citep{wei2022chain,wei2022emergent}, and step-by-step planning \citep{huang2022language,wang2023describe}.
In pursuit of advancing the capabilities of LLMs for practical applications, an ongoing research direction is investigating the incorporation of external tools/APIs to enhance the functionality of LLMs\citep{mialon2023augmented,qin2023tool,parisi2022talm,schick2023toolformer}. This endeavor has yielded successful integration of diverse tools, including search engines and other foundational models, with LLMs\citep{lu2023chameleon,shen2023hugginggpt,wu2023visual}.

Despite significant progresses, we find that existing API-augmented LLMs are still in the experimental stage and have yet to fully meet the demands of real-world user instructions.
As shown in Table \ref{tab:comparison}, current methods are limited to connect with a small number of specially designed tools/APIs \citep{schick2023toolformer,lu2023chameleon,yang2023gpt4tools}.
For example, Chameleon \citep{lu2023chameleon} designs a set of 15 tools, such as table verbalizer and image captioner.
Additionally, the absence of a standardized API design specification obstructs the scalability of previous endeavors.
Thus, the potential for connecting LLMs with a diverse range of real-world APIs, like RESTful APIs, remains under-explored and challenging.
Furthermore, when dealing with a complex instruction in real scenario, it is necessary to decompose it into smaller sub-tasks and accomplish them by employing a mix of various APIs.
As a result, it becomes essential for API-augmented LLMs to have robust planning and decision-making capabilities to effectively tackle real-world tasks \citep{qin2023tool}.
Nonetheless, existing techniques, either offline introspective plan-then-execute methods \citep{qin2023tool,shen2023hugginggpt,lu2023chameleon}  or the ReAct framework \citep{yao2022react}, encounter challenges in effectively adapting API feedback and generating viable plans.

\begin{table*}[t!]
\centering
\resizebox{1.0\linewidth}{!}{
\begin{tabular}{l|ccc|cccc}
\toprule
\multicolumn{1}{l}{\multirow{2}{*}{Model}} & \multicolumn{3}{c}{API/Tool Use} & \multicolumn{4}{c}{Framework} \\
\cmidrule(lr){2-4} \cmidrule(lr){5-8}
\multicolumn{1}{c}{} & Num. & Extensibility & Schema & Planning & Planning Form & Feedback & Plug-n-Play \\
\midrule
ReAct & 3 & $-$ & Specialized & Online & Natural Lang. & \cmark & \cmark \\
Toolformer & 5 & $-$ & Specialized & \xmark & - & \xmark & \xmark \\
Visual ChatGPT & 22 & $-$ & Specialized & \xmark & - & Human & \cmark \\
ViperGPT & 11 & $-$ & Python func. & Offline & Program & \xmark & \cmark \\
HuggingGPT & 24\footnotemark{} & $+$ & HuggingFace & Offline & Natural Lang. & \xmark & \cmark \\
API-Bank & 53 & $-$ & Specialized & \xmark & - & Human & \cmark \\
Chameleon & 15 & $-$ & Specialized & Offline & Natural Lang. & \xmark & \cmark \\
Gorilla & 1645$^{\dagger}$ & $+$ & JSON & \xmark & - & \xmark & \xmark \\
GPT4Tools & 31 & $-$ & Specialized & \xmark & - & Human & \xmark \\
\midrule
RestGPT (\textbf{ours}) & 100+ & $++$ & RESTful & Online & Coarse-to-Fine & \cmark & \cmark \\
\bottomrule
\end{tabular}
}
\caption{A comparison of work that augments LLMs with API/tool usage. $^{\dagger}$ denotes API selection with retrieval.}
 \label{tab:comparison}
\end{table*}
\footnotetext{HuggingGPT \citep{shen2023hugginggpt} claims it has integrated hundreds of models on HuggingFace. However, all of the models only cover 24 tasks such as text classification, object detection, etc.}

In this work, we delve into a more realistic scenario by connecting LLMs with real-world RESTful APIs, aiming at fulfilling practical user instructions.
RESTful is the \textit{de facto} standard for web service development \citep{li2016rest}, which utilizes HTTP methods (e.g., \texttt{GET}, \texttt{POST}) and URIs to manipulate resources.
RESTful API development typically adheres to the OpenAPI Specification (OAS) \citep{swagger}, which describes the operations, parameters, and response schemas of each API endpoint.
Therefore, our resulting framework can  connect with any RESTful application and offer standardized API development processes, thereby enabling enhanced extensibility compared to previous approaches.
However, connecting LLMs with RESTful APIs also brings practical challenges. 
First, calling real-world APIs may give rise to a multitude of unforeseen situations, necessitating the framework to exhibit strong robustness and conduct reasonable planning.
Second, the parameters and responses of RESTful APIs often follow specific formats, leading to difficulty in API invoking and response parsing.

To tackle the limitations of previous methods and the practical challenges associated with RESTful APIs, we propose \ourmethod{}, a LLM-based framework connecting with RESTful APIs to handle complex instructions.
\ourmethod{} comprises three main modules: a Planner, an API Selector, and an Executor.
The core of each module is prompting an LLM.
Unlike prior work that uses static or ReAct style planning which lacks flexibility in realistic scenarios, \ourmethod{} adopts an iterative coarse-to-fine online planning mechanism.
Given a complicated instruction, the planner generates a sub-task for current task in the format of natural language.
Subsequently, the API selector maps the coarse high-level sub-task to finer API calling plan, forming a coarse-to-fine task planning.
The executor, responsible for invoking RESTful APIs and get execution results, is further divided it into two sub-modules: a Caller and a response Parser. 
The caller organizes API call parameters based on the API plan and API documentation, while the parser utilizes the response schema defined in OAS to generate Python code to parse responses.
Once receiving the execution results of the API plan, the planner performs online planning for the subsequent sub-task in the next step.
Through the integration of the three modules, our method \ourmethod{} shows superior extensibility and flexibility in mastering RESTful APIs.

To evaluate the performance of \ourmethod{} in utilizing RESTful APIs, we introduce RestBench, a human-annotated benchmark consisting of two realistic scenarios, TMDB movie database and Spotify music player.
For each scenario, we collect diverse real-world user instructions that require the utilization of multiple APIs to complete.
Based on the RestBench, we conduct comprehensive experiments to investigate the performance of \ourmethod{} across different dimensions.
The experimental results demonstrate that RestGPT exhibits robust capabilities in handling complex user instructions and has significant advantages in task planning, API understanding, and response parsing.

Our contributions can be summarized as follows:
\begin{enumerate}
    \item For the first time, we attempt to connect large language models with RESTful APIs, enabling the resulting framework to be compatible with existing real-world applications while also providing powerful extensibility.
    \item We propose \ourmethod{}, a coarse-to-fine online planning framework that effectively handles the practical challenges associated with connecting LLMs with RESTful APIs, including API understanding, planning, and API response parsing.
    \item To evaluate the performance of \ourmethod{}, we build a human-annotated benchmark, RestBench, which comprises two practical scenarios.
    Experimental results show the capability of \ourmethod{} to effectively utilize a number of RESTful APIs to accomplish complex instructions.
\end{enumerate}

\section{Background}

\subsection{Tool-Augmented Language Models}

The emergence of recent powerful LLMs has enabled artificial intelligence systems to match human skills in utilizing tools \citep{mialon2023augmented,qin2023tool}.
To enhance the performance of LLMs in accessing up-to-date information and carrying out precise mathematical reasoning, early work leverages simple tools like web search engines and calculators, such as ReAct \citep{yao2022react}, Toolformer \citep{schick2023toolformer}, and ART \citep{paranjape2023art}.
Another line of research has focused on equipping LLMs to coordinate with external models for complex AI tasks, exemplified by HuggingGPT \citep{shen2023hugginggpt}, ViperGPT \citep{suris2023vipergpt}, Visual ChatGPT \citep{wu2023visual} and Chameleon~\citep{lu2023chameleon}. 
Recently, some work study how to enable open-sourced LLMs, such as LLaMa, to perform API usage \citep{patil2023gorilla,yang2023gpt4tools,tang2023toolalpaca}.
Additionally, API-Bank~\citep{li2023apibank} provides a systematic benchmark to showcase the efficacy of LLMs using tools to respond to human instructions. 

Despite the notable advancements in incorporating tools for large language models, previous methods have exhibited certain limitations, most notably their restricted support for a limited number of specially designed APIs \citep{lu2023chameleon} and their inferior planning methods \citep{qin2023tool,wang2023large,lu2023chameleon}. 
We compare \ourmethod{} with other tool-augmented language models in Table \ref{tab:comparison}.
As shown, our work stands out by supporting for over 100  RESTful APIs.
Furthermore, compared with most previous approaches adopt static offline planning which cannot interact with APIs and utilize feedback to adjust the plan, we employ a coarse-to-fine online planning framework with feedback, facilitating more flexible planning for complex instructions.
Our work shares the similar spirit of AutoGPT, an autonomous agent capable of accomplishing complex tasks with numerous tools.
While AutoGPT relies on developers to ensure compatibility with various applications, \ourmethod{} can be integrated with any RESTful API-based applications in a plug-and-play fashion.

\subsection{RESTful APIs}

RESTful APIs have become a popular way to expose functionalities and data of web services to client applications \citep{masse2011rest,li2016rest}. 
RESTful APIs also provide a standard for integrating external systems together with using a simple yet powerful interface.
There are millions of RESTful APIs available on Internet, such as Spotify, Twitter, Gmail, etc.
RESTful APIs are based on the REST architectural style, which emphasizes a client-server communication via stateless HTTP requests, including \texttt{GET}, \texttt{POST}, etc, where resources are identified by self-descriptive URIs \citep{masse2011rest}.
The response of RESTful APIs are always structured in JSON format and contain various information.
Thus, LLMs connected with RESTful APIs must possess a strong ability to extract the required information from the response.

OpenAPI Specification (OAS, or Swagger) \citep{swagger}, has been widely adopted as a standard for defining RESTful APIs.
OAS is a structured documentation file which describes the endpoints, operations, parameters, response schemas, and other details of an API endpoint, providing a clear interface for our method to use the APIs.

\section{RestGPT}

\begin{figure*}[t!]
    \centering
    \includegraphics[width=0.95\linewidth]{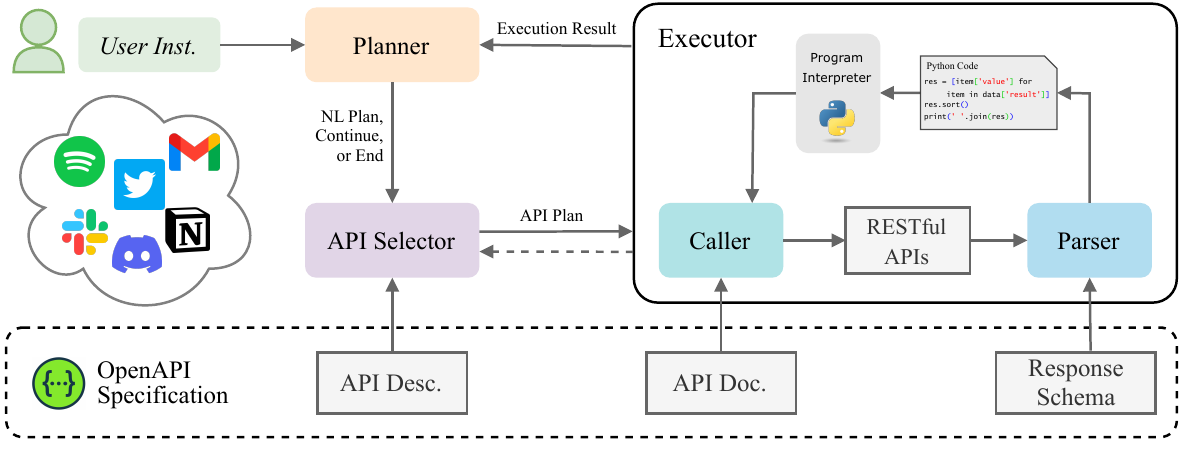}
    \caption{
    Overview of \ourmethod{}. 
    The planner, API selector, executor collaborate to form the coarse-to-fine online planning framework.
    The caller and response parser in the executor provides robust execution of the RESTful API calling plan.
    }
    \label{fig:model}
\end{figure*}

\subsection{\ourmethod{} Architecture}

As demonstrated in Figure \ref{fig:model},
\ourmethod{} is composed of three main modules: a \textbf{Planner} $\mathcal{P}$, an \textbf{API Selector} $\mathcal{S}$ and an \textbf{Executor} $\mathcal{E}$.
The planner decomposes each user instruction into several sub-tasks, while the API selector selects APIs to address each sub-task.
The executor, consisting of a \textbf{Caller} and a response \textbf{Parser}, performs RESTful API calls and extracts useful information from the JSON response to form the execution result.
The core of each component is an LLM with the corresponding prompt and in-context examples describing the function of the component.

One of the challenges in connecting LLMs with a vast number of APIs is to ensure that the framework is able to fully understand the API documents with a limited context window size of LLMs.
As depicted in Figure \ref{fig:model}, we designate different modules to read distinct parts of the OpenAPI Specification (OAS). 
This strategy allows us to leverage OAS information to its fullest potentials when working with RESTful APIs.
Specifically, the API selector reads the endpoint descriptions of all APIs to select a proper API for solving the current sub-task. 
Then, the caller uses the detailed documents of the API within the API plan to generate the correct API calling parameters and request body. 
Lastly, the parser is developed to make use of the response schema within OAS to generate the parsing code for information extraction.

\subsection{Coarse-to-fine Online Planning}

To fully exploit the planning and decision making capabilities of LLMs and enable our method to dynamically adjust the plan to changing circumstances when accomplishing real-world user instructions, we propose a coarse-to-fine online planning mechanism in \ourmethod{}.

The workflow of \ourmethod{} can be characterized as an iterative ``plan and execution'' loop.
During the planning stage, the planner and API selector collaborate to accomplish  an instruction through iteratively decomposing it into suitable natural language sub-tasks and corresponding APIs.
In each step $t$, the planner $\mathcal{P}$ leverages commonsense knowledge to generate a natural language (NL) sub-task $p_t$ based on the user instruction $q$, previous NL plans $(p_1, ..., p_{t-1})$, and execution results $(r_1, ..., r_{t-1})$, thereby constructing a high-level NL plan.
Then, the API selector $\mathcal{S}$ reads the descriptions of available API endpoints to select appropriate APIs and construct the finer API plan $a_t$, which may contain a single or multiple API calls to solve the current NL plan $p_t$.
Then the executor $\mathcal{E}$ executes the API plan $a_t$ and gets the execution result $r_t$ for current step.
This process can be formulated as:
\begin{equation}
\begin{aligned}
\text{NL Plan: } & p_t\leftarrow \mathcal{P}(q;p_{1}, r_{1} ..., p_{t-1}, r_{t-1}), \\
\text{API Plan: } & a_t\leftarrow \mathcal{S}(p_t;r_1, ..., r_{t-1}), \\
\text{Exec. Res.: } & r_t\leftarrow \mathcal{E}(a_t;r_1, ..., r_{t-1}).
\end{aligned}
\end{equation}
In this way, the planner and API selector are dedicated to NL sub-task planning and API selection, respectively, effectively utilizing the large language model's abilities of planning and text comprehension.

Alongside the ``plan and execution'' loop, we design two special states, ``continual'' and ``end'', for the planner to monitor the execution result from the executor.
Specifically, if the planner finds that the current executor's output $r_t$ has not completed the present NL sub-task $p_t$, it will output a ``continue'' signal and provide
a special NL plan $p_{t+1}$
to the API selector, instructing it to continue fulfilling the plan $p_t$.
In such cases, the API selector will re-generate a new API plan based on the original NL plan $p_t$, new NL plan $p_{t+1}$, previous API plan $a_t$ and execution result $r_t$.
This process is described as:
\begin{equation}
\begin{aligned}
\text{API Plan: } & a_{t+1}\leftarrow \mathcal{S}(p_t,p_{t+1};r_1, ..., r_{t-1};a_t,r_t), \\
\text{Exec. Res.: } & r_{t+1}\leftarrow \mathcal{E}(a_{t+1};r_1, ..., r_{t-1},r_t).
\end{aligned}
\end{equation}
If the planner assesses that the user's request has been completed, it will give the termination signal ``end'' and output the final result.
With such a design, our method achieves a more flexible online planning which is capable of handling various situations encountered in real-world scenarios.

The planner, API selector, and executor collaborate to form RestGPT's coarse-to-fine online planning framework.
This framework significantly enhances the ability to decompose tasks and select appropriate APIs, providing the model with the flexibility to effectively tackle user instructions.

\subsection{API Plan Execution}
\label{sec:executor}

\begin{wrapfigure}{r}{0.49\textwidth}
    \centering
    \includegraphics[width=1.0\linewidth]{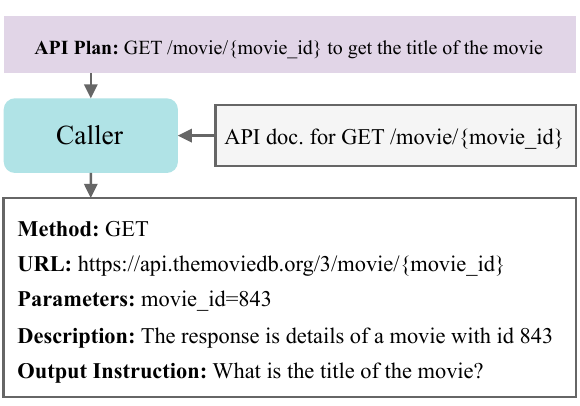}
    \caption{
        Example output of the caller.
    }
    \label{fig:caller}
\end{wrapfigure}

Once an API calling plan is generated, the next step is to execute it.
The executor $\mathcal{E}$ consists of a caller and a response parser.
The caller should read the API documents carefully and generate correct parameters or request body for the API call.
Due to the constraints of maximum context length, we filter API documents and only preserve APIs appearing in current API plan $a_t$.
Given the generated parameters and request body, we use Requests Python library to call the RESTful API.
Besides, to guide the response parser to extract information from the API response, the caller also generates a response description and output instruction for the response parser.
Figure \ref{fig:caller} presents an example output of the caller.

RESTful APIs typically return a JSON formatted response with much redundant information. 
The executor needs to extract the required information from the response and return it to the planner.
However, the response may sometimes have a complex structure or be lengthy, making it difficult to extract important information via directly prompting the LLMs.
To address this problem, we make use of the response schema defined in the OAS. 
Specifically, we utilize the coding capability of LLM to generate Python parsing code based on the provided schema and output instructions generated by the caller. 
Next, the Python code is executed to get the final result. 
If there are no execution exceptions or errors, the output is returned. 
Otherwise, the LLM is prompted to parse the response directly as a backup.

\section{RestBench}

To assess the effectiveness of \ourmethod{} in processing complex user instructions through RESTful APIs, we introduce RestBench, a high-quality human annotated dataset comprising of two real-world scenarios.
Existing researches have proposed several benchmarks for the evaluation of tool/API augmented LLMs \citep{li2023apibank,patil2023gorilla,qin2023tool}. 
However, these benchmarks primarily focus on simple tasks that can be accomplished using a single API.
We hope RestBench can facilitate the exploration on utilizing multiple APIs to address real-world user instructions.

\subsection{Scenarios and APIs}

We select two common real-world scenarios: TMDB movie database and Spotify music player.
The main consideration is to evaluate the capabilities of RestGPT: (1) augmenting LLMs with external specialized domain database via RESTful APIs; (2) connecting LLMs with RESTful APIs to autonomously control real-world applications.
TMDB offers official RESTful APIs encompassing the information of movies, TVs, actors, and images. 
Spotify music player provides API endpoints to retrieve content metadata, receive recommendations, create and manage playlists, and control playback.
For these two scenarios, we filter out 54 and 40 commonly used APIs respectively and obtain the corresponding OpenAPI Specifications to build RestBench.

\subsection{Dataset Collection}

\begin{table}[t!]
\centering
\scalebox{0.97}{
\begin{tabular}{lccccccc}
\toprule
\multicolumn{1}{l}{\multirow{2}{*}{Scenario}} & \multicolumn{1}{c}{\multirow{2}{*}{\makecell[c]{Num. \\APIs}}} & \multicolumn{4}{c}{Len. of Solution Path} & \multicolumn{1}{c}{\multirow{2}{*}{\makecell[c]{Avg. \\Len.}}} & \multicolumn{1}{c}{\multirow{2}{*}{Total}} \\
\cmidrule(lr){3-6}
\multicolumn{1}{c}{} & \multicolumn{1}{c}{} & 1 & 2 & 3 & 4 & \multicolumn{1}{c}{} & \multicolumn{1}{c}{} \\
\midrule
TMDB & 54 & 5 & 66 & 27 & 2 & 2.3 & 100 \\
Spotify & 40 & 8 & 18 & 22 & 9 & 2.6 & 57 \\
\bottomrule
\end{tabular}
}
\vspace{2.0mm}
\caption{
Statistics of RestBench test set.
We report the number of instructions with different lengths of solution path.
}
 \label{tab:dataset_stat}
\end{table}

\begin{wraptable}{r}{0.49\textwidth}
\centering
\small
\scalebox{0.95}{
\begin{tabular}{l}
\toprule
\mybox{
\centering
$\triangleright$ \textit{TMDB} 
} \\
\textbf{Instruction:} \\
Who is the director of today's most trending movie? \\
\\
\textbf{Gold Solution Path:} \\
1. \texttt{GET /trending/\{media\_type\}/\{time\_window\}} \\
2. \texttt{GET /movie/\{movie\_id\}/credits} \\
\midrule
\mybox{
\centering
$\triangleright$ \textit{Spotify} 
} \\
\textbf{Instruction:} \\
\makecell[l]{Make me a playlist containing three songs of Mariah \\Carey and name it 'Love Mariah'} \\
\\
\textbf{Gold Solution Path:} \\
1. \texttt{GET /search} \\
2. \texttt{GET /me} \\
3. \texttt{POST /users/\{user\_id\}/playlists} \\
4. \texttt{POST /playlists/\{playlist\_id\}/tracks} \\
\bottomrule
\end{tabular}
}
\caption{
Example instructions and the corresponding gold solution paths of RestBench.
}
 \label{tab:dataset_example}
    \vspace{-2mm}
\end{wraptable}

High-quality instructions generally satisfy two crucial aspects: (1) to reflect a wide range of real user needs;  (2) to cover different levels of complexity to fully study the reasoning and planning ability of our method.
To achieve these goals, we adopt a bottom-up instruction collection approach.
We employ 6 experts that work on NLP research to brainstorm instructions for different combinations of APIs.
Along with the instructions, the experts need to annotate the gold API solution path for each instruction.
To guarantee the quality of the instructions, we employ two additional experts to thoroughly verify the solvability of each instruction and correctness of the corresponding solution path.
Ultimately, we annotate 10 instruction-solution pairs for each scenario as the development set, and 100 pairs for TMDB and 57 pairs for Spotify as the test set.
Though the data scale is not large, these instructions are typical of the frequently raised user requests.
Moreover, different from prior work which uses LLMs to get API calling procedure, we utilize human labeled API solution paths for evaluation.
Table \ref{tab:dataset_example} presents example instructions of the two scenarios.
The statistics of RestBench are shown in Table \ref{tab:dataset_stat}.

\subsection{Evaluation Metrics}

Since some user requests are time-dependent (see the TMDB example in Table \ref{tab:dataset_example}),
it is impractical to annotate a fixed ground-truth answer for each instruction, whereas, the API solution paths for most instructions remain consistent.
If the model-generated API call path contains the gold API call path as a subsequence (with the elements not necessarily being contiguous), we think that the model has generated a correct path.
To further evaluate the model's performance, we rely on human evaluation to determine if the model result successfully fulfills the user query.
We calculate the proportion of correct paths and successful query completions as metrics, i.e., \textbf{Correct Path Rate} and \textbf{Success Rate}. 
Moreover, the number of actual API calls can be utilized to measure the planning efficiency of different methods.
Given the length of gold solutions, we further define  \textbf{$\Delta$ Solution Len.} as the mean number of additional API calls required to successfully execute an instruction:
$$
\Delta \text{Solution Len.}=\frac{1}{N_s}\sum_{i=0}^N(L^i_{real}-L^i_{gold})\cdot \mathbb{I}(i, \text{success}),
$$
where $N_s$ is the number of successfully accomplished instructions, $L^i_{real}$ and $L^i_{gold}$ are the actually and gold number of API calls for the $i$-th instruction respectively, $\mathbb{I}(i, \text{success})$ denotes whether the $i$-th instruction is successfully completed.

\section{Experiments}

\subsection{Experimental Setup}

We compare \ourmethod{} with four recent baselines, including offline introspective method \citep{qin2023tool} used in HuggingGPT \citep{shen2023hugginggpt} and Chameleon \citep{lu2023chameleon}, DEPS \citep{wang2023describe}, ReAct \citep{yao2022react} and Reflexion \citep{shinn2023reflexion}.
Since some methods are not originally designed for tool/API usage, we reproduce them and add the API executor proposed in Section \ref{sec:executor} to make them able to call RESTful APIs.
The maximum steps for DEPS is set to 10 and the maximum trials for Reflexion is set to 2.

To showcase the planning and API calling capabilities of our method, we implement two ablation variants of \ourmethod{}.
The first variant involves removing the planner and allowing the API selector to directly choose APIs in a ReAct style. 
This approach can be seen as ReAct equipped with our proposed executor.
The second one is to replace the schema-based response parser with an LLM that directly reads and extracts the required information from the JSON response.

In our experiments, we employ text-davinci-003 from OpenAI as the LLM for \ourmethod{} and all baselines.
The decoding temperature is set to 0 for the most deterministic generation.

\begin{table*}[t!]
\centering
\resizebox{1.0\linewidth}{!}{
\begin{tabular}{l|ccc|ccc}
\toprule
\multicolumn{1}{l}{\multirow{2}{*}{Model}} & \multicolumn{3}{c}{TMDB} & \multicolumn{3}{c}{Spotify} \\
\cmidrule(lr){2-4} \cmidrule(lr){5-7}
\multicolumn{1}{c}{} & Success\% & CP\% & $\Delta$ Solution Len. & Success\% & CP\% & $\Delta$ Solution Len. \\
\midrule
Offline \citep{qin2023tool} & 29.0 & 33.0 & +1.52 & 14.5 & 36.4 & +1.10 \\
DEPS \citep{wang2023describe} & 38.0 & 43.0 & +1.20 & 19.3 & 43.8 & +1.74 \\
ReAct \citep{yao2022react} & 44.0 & 57.0 & +0.76 & 54.5 & 49.1 & +0.31 \\
Reflexion \citep{shinn2023reflexion} & 52.0 & 59.0 & +1.37 & 59.6 & 61.4 & +1.68 \\
\midrule
RestGPT & \textbf{75.0} & \textbf{79.0} & \textbf{+0.55} & \textbf{72.7} & \textbf{74.5} & +0.25 \\
\quad w/o Planner$^\dagger$ & 44.0 & 57.0 & +0.76 & 54.5 & 49.1 & +0.31 \\
\quad w/o Parser & 46.0 & 53.0 & +0.60 & 47.3 & 52.7 & \textbf{+0.24} \\
\midrule
\ourmethod{} (ChatGPT) & 68.0 & 65.0 & +0.72 & 69.1 & 72.3 & +0.28 \\
\ourmethod{} (Llama2-13B) & 0.0 & 0.0 & - & 0.0 & 0.0 & - \\
\ourmethod{} (Vicuna-13B) & 9.0 & 15.0 & +1.21 & 12.7 & 20.6 & +1.52 \\
\bottomrule
\end{tabular}
}
\caption{
Success rate (\%), Correct Path rate (CP, \%), and $\Delta$ Solution Length on two scenarios of RestBench. 
The best results are in \textbf{boldface}.
$\dagger$ \ourmethod{} w/o planner is equivalent with ReAct equipped with our proposed executor.
}
 \label{tab:main_exp}
\end{table*}

\subsection{Main Results}
\label{sec:main_res}

Table \ref{tab:main_exp} shows the performance of \ourmethod{} and baselines on two scenarios.
Our approach outperforms all other methods in both scenarios, achieving a success rate of 75\% on the movie database and over 70\% on the music player.
Note that in most cases, the correct path rate is slightly higher than success rate, indicating that the method may generate  correct API calling plan but fail to execute it.
\ourmethod{} also stands out with its minimal solution length, showcasing the superior planning ability of the coarse-to-fine online planning mechanism.

Ablation experiments on coarse-to-fine planning and schema-based parser show both mechanisms are conductive to the model performance.
Particularly, when removing the planner, the performance degrades significantly, indicating that current LLMs are unable to simultaneously conduct planning, API understanding and selection.
Thus, the coarse-to-fine planning mechanism plays a crucial role in our framework.
The ablation results without parser demonstrates that the schema-based parser enables LLMs to better comprehend and parse the real-world API responses with complicated structure.

To investigate the performance of our method with different base LLMs, we implement \ourmethod{} with ChatGPT (gpt-3.5-turbo-0301), Llama2-13B (Llama-2-13b-chat-hf), and Vicuna-13B (vicuna-13b-v1.5).
As shown in Table \ref{tab:main_exp}, the performance of ChatGPT is slightly worse than text-davinci-003.
Interestingly, we have tried all official checkpoints of Llama2-13B, but none of them were able to comprehend the prompt and generate valid plans.
In contrast, Vicuna-13B, which is fine-tuned from Llama2 on user-shared conversations, can accomplish some simple instructions.
This result indicates that by fine-tuning LLMs on ChatGPT-generated data, the model can acquire the ability to understand and follow complicate prompts.

\begin{figure*}[t]
    \centering
    \includegraphics[width=0.9\linewidth]{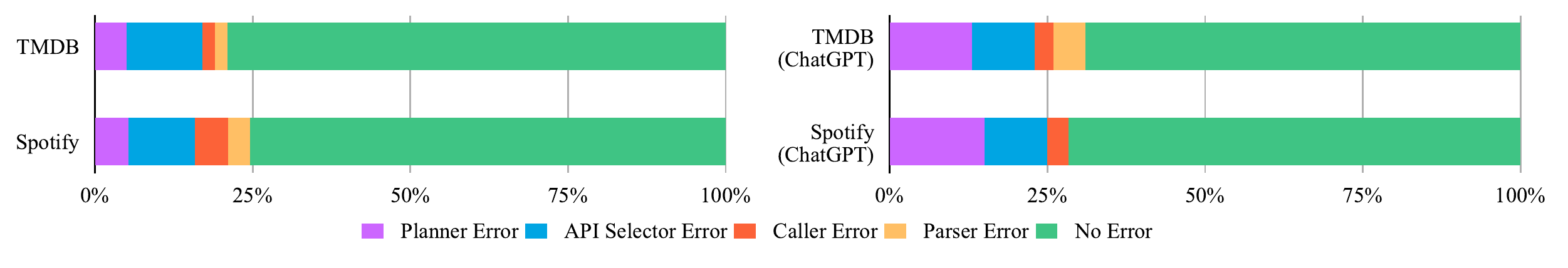}
    \caption{
        Error breakdown of \ourmethod{} on RestBench.
        Error types are categorized by the module where the error occurred.
    }
    \label{fig:error_analysis}
\end{figure*}

\begin{figure*}[t]
    \centering
    \includegraphics[width=0.95\linewidth]{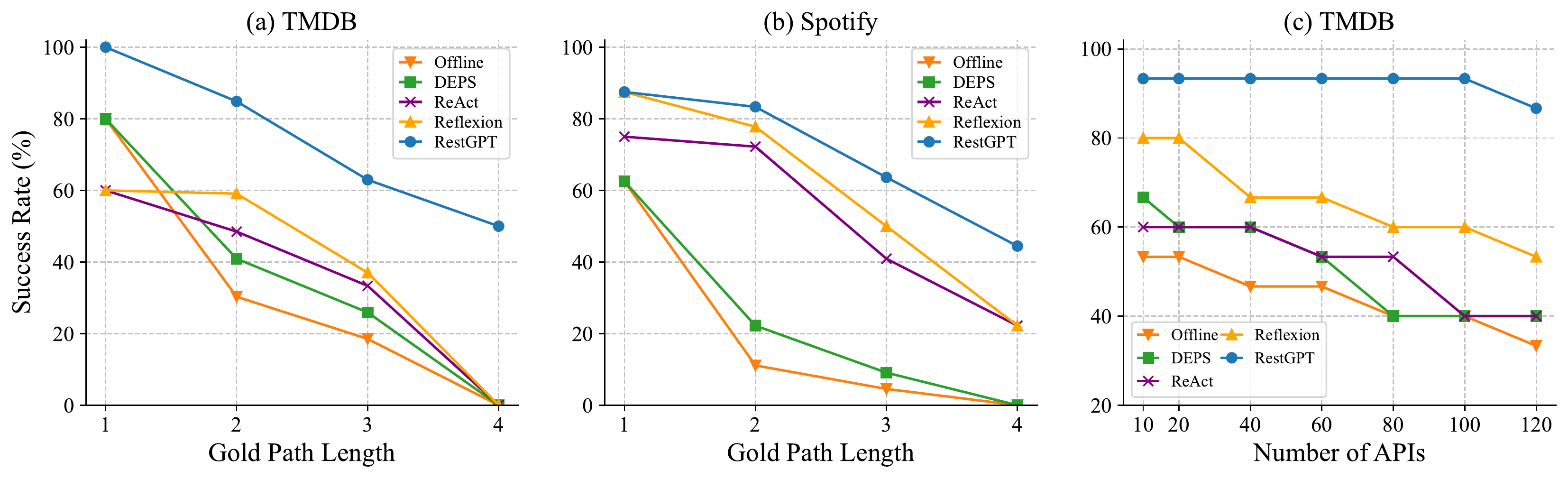}
    \caption{
        Scaling ability of \ourmethod{}.
        (a) (b) Scaling curves of the gold solution path on TMDB and Spotify.
        The length of gold API solution path indicates the complexity of the instruction.
        (c) Scaling curves of the number of APIs on TMDB scenario.
    }
    \label{fig:scaling_path}
\end{figure*}

\subsection{Error Analysis}
\label{sec:error}

To further investigate the effectiveness of different modules in \ourmethod{}, we conduct error analysis.
In Figure \ref{fig:error_analysis}, we classify errors based on the module in which they occur.
We discover that the majority of errors occur during the planning stage, i.e., within the planner (purple) and API selector (blue).
The planner sometimes loses track of its intended objective after multiple rounds of execution, resulting in early exit before completing the instruction.
For the API selector, it may either select incorrect APIs or hallucinate to make up in-path parameters.
This error analysis highlights the insufficient planning and decision-making capabilities of LLMs.

Compared with text-davinci-003, ChatGPT tends to make more errors in the planning stage, leading to slightly worse performance on both scenarios.
More specifically, we find that ChatGPT is often too verbose and tend to continue planning even after the user instruction has been fulfilled.
This behavior can be attributed to the fact that ChatGPT is trained specifically for conversational interactions, which encourages it to generate more lengthy responses.

\subsection{Scaling Curves}

In this section, we aim to demonstrate the scaling ability of \ourmethod{} on two dimensions: scaling the difficulty of the tasks and scaling the number of APIs.

For each instruction in RestBench, the length of gold solution path indicates the complexity of the instruction.
We calculate the success rate of models on instructions with varying complexities.
As depicted in Figure \ref{fig:scaling_path} (a) (b), the success rate of all methods decreases as the complexity of the instruction increases.
Notably, when the gold path length is 4, all baselines struggle to complete the task in both scenarios. In contrast, our proposed \ourmethod{} can still achieve a success rate of over 40\%, showing its superior performance in planning and API calling.

Before conducting experiments on scaling the number of APIs, we handpicked 10 APIs from TMDB and created a small test set comprising 15 instructions.
All 15 instructions can be resolved using the selected 10 APIs.
Then, we increasingly expanded the number of APIs and introduced additional noise APIs sourced from the official TMDB APIs.
The results are shown in Figure \ref{fig:scaling_path} (c).
As the number of noise APIs increases, the performance of all baseline methods deteriorates due to their inferior planning and reasoning.
However, our method almost remains unaffected. 
These results effectively demonstrate the strong extensibility of our proposed \ourmethod{}.

\subsection{Case Study}

\begin{figure*}[t!]
    \centering
    \includegraphics[width=1.0\linewidth]{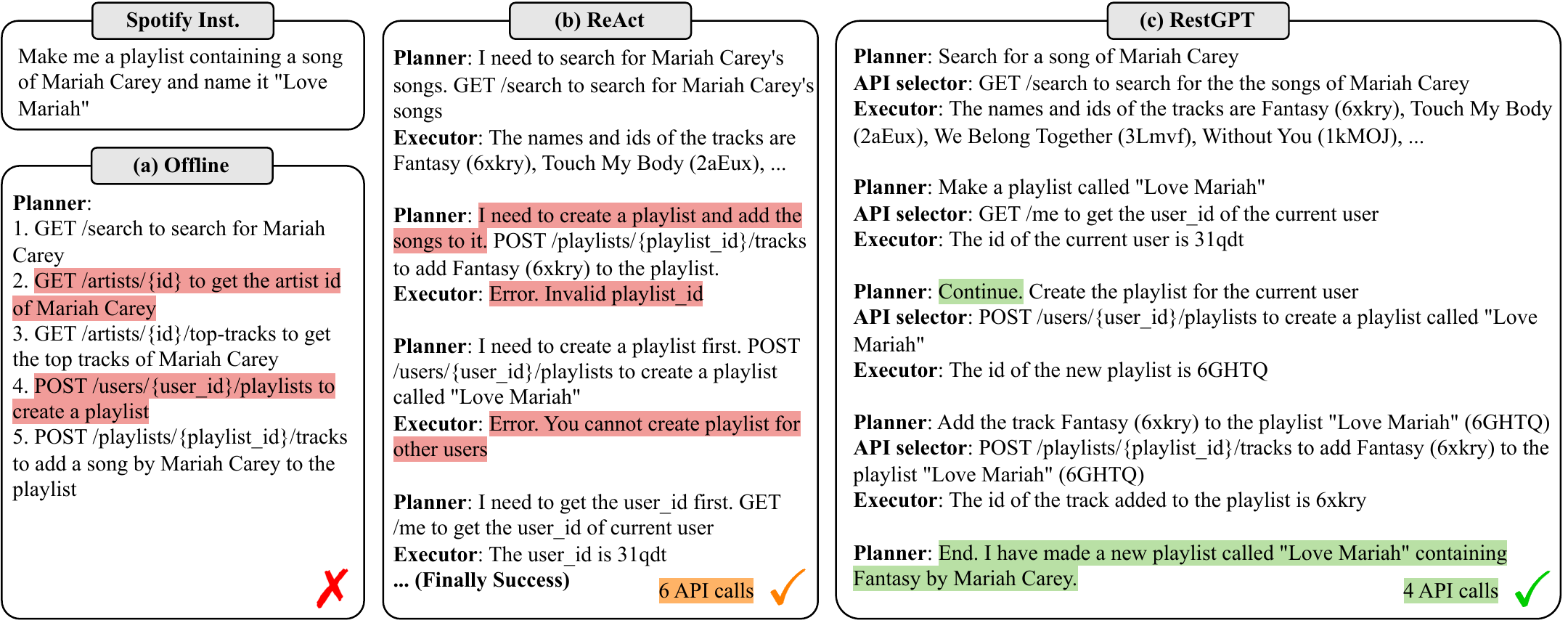}
    \caption{
Case study of three methods, (a) Offline \citep{qin2023tool,shen2023hugginggpt,lu2023chameleon}, (b) ReAct \citep{yao2022react}, and (c) \ourmethod{}.
For offline method, we only show the generated plan.
For ReAct and \ourmethod{}, we omit the detailed execution process of the executor.
    }
    \label{fig:compare_case}
\end{figure*}

In Figure \ref{fig:compare_case}, we conduct a case study to compare the planning ability of \ourmethod{} with the offline planning \citep{qin2023tool,lu2023chameleon} and ReAct \citep{yao2022react} framework.
Firstly, we observe the offline method is unable to solve most user instructions.
As depicted in Figure \ref{fig:compare_case} (a), the planner not only selects the wrong API (step 2), but also ignores the dependencies between APIs and used the parameter ``user\_id'' before obtaining it (step 4).
Regarding ReAct which generates chain-of-thought and actions in an interleaved manner, we find that current LLMs have a limited ability to simultaneously conduct planning, API understanding and selection.
As shown in Figure \ref{fig:compare_case} (b), the planner of ReAct generates a sub-task that is difficult to solve (step 2) and also ignores the dependencies between different APIs (step 3).
Due to the inferior planning, it consumes 6 API calls to complete the task.
In contrast, \ourmethod{} employs a planner to generate high-level NL sub-tasks and an API selector to choose appropriate APIs to solve the sub-task.
Notably, in step 3, the planner assesses the playlist that has not been successfully created and generate "continue" signal with further instructions for the API selector.
Our method accomplishes the instruction with only 4 API calls.
The coarse-to-fine online planning framework of \ourmethod{} fully exploits the LLMs' planning and document understanding capabilities, providing the model with the flexibility to tackle complex user requests.

\section{Conclusion}

In this paper, we explore the scenarios of connecting current large language models (LLMs) with real-world applications via RESTful APIs.
To overcome the limitations of existing approaches and tackle the challenges in integrating LLMs with RESTful APIs, we propose \ourmethod{}, an approach that leverages LLMs to complete complex user instructions. 
Our method features a coarse-to-fine online planning mechanism to enable more flexible planning and API selection. 
Furthermore, to handle the complex scenario of calling RESTful APIs, we designed a specialized API executor to formulate parameters and parse API responses. 
To assess the performance of our method, we build a high-quality dataset, RestBench, which consists of human-annotated instructions from two realistic scenarios.
Extensive experiments demonstrate that \ourmethod{} achieves impressive results in complex tasks and exhibits strong robustness, which paves a new way towards AGI.
In the future, we aim to delve into a broader range of intricate tasks, thoroughly examining the immense potential of \ourmethod{} across both academic and industrial domains.

{\small
\bibliographystyle{unsrtnat}
\bibliography{ref}

\begin{thebibliography}{26}
\providecommand{\natexlab}[1]{#1}
\providecommand{\url}[1]{\texttt{#1}}
\expandafter\ifx\csname urlstyle\endcsname\relax
  \providecommand{\doi}[1]{doi: #1}\else
  \providecommand{\doi}{doi: \begingroup \urlstyle{rm}\Url}\fi

\bibitem[Brown et~al.(2020)Brown, Mann, Ryder, Subbiah, Kaplan, Dhariwal,
  Neelakantan, Shyam, Sastry, Askell, et~al.]{brown2020language}
Tom Brown, Benjamin Mann, Nick Ryder, Melanie Subbiah, Jared~D Kaplan, Prafulla
  Dhariwal, Arvind Neelakantan, Pranav Shyam, Girish Sastry, Amanda Askell,
  et~al.
\newblock Language models are few-shot learners.
\newblock \emph{Advances in neural information processing systems},
  33:\penalty0 1877--1901, 2020.

\bibitem[OpenAI(2022)]{openai2022chatgpt}
OpenAI.
\newblock Chatgpt, 2022.
\newblock URL \url{https://openai.com/blog/chatgpt}.

\bibitem[Dong et~al.(2022)Dong, Li, Dai, Zheng, Wu, Chang, Sun, Xu, and
  Sui]{dong2022survey}
Qingxiu Dong, Lei Li, Damai Dai, Ce~Zheng, Zhiyong Wu, Baobao Chang, Xu~Sun,
  Jingjing Xu, and Zhifang Sui.
\newblock A survey for in-context learning.
\newblock \emph{arXiv preprint arXiv:2301.00234}, 2022.

\bibitem[Wei et~al.(2022{\natexlab{a}})Wei, Wang, Schuurmans, Bosma, Chi, Le,
  and Zhou]{wei2022chain}
Jason Wei, Xuezhi Wang, Dale Schuurmans, Maarten Bosma, Ed~Chi, Quoc Le, and
  Denny Zhou.
\newblock Chain of thought prompting elicits reasoning in large language
  models.
\newblock \emph{arXiv preprint arXiv:2201.11903}, 2022{\natexlab{a}}.

\bibitem[Wei et~al.(2022{\natexlab{b}})Wei, Tay, Bommasani, Raffel, Zoph,
  Borgeaud, Yogatama, Bosma, Zhou, Metzler, et~al.]{wei2022emergent}
Jason Wei, Yi~Tay, Rishi Bommasani, Colin Raffel, Barret Zoph, Sebastian
  Borgeaud, Dani Yogatama, Maarten Bosma, Denny Zhou, Donald Metzler, et~al.
\newblock Emergent abilities of large language models.
\newblock \emph{arXiv preprint arXiv:2206.07682}, 2022{\natexlab{b}}.

\bibitem[Huang et~al.(2022)Huang, Abbeel, Pathak, and
  Mordatch]{huang2022language}
Wenlong Huang, Pieter Abbeel, Deepak Pathak, and Igor Mordatch.
\newblock Language models as zero-shot planners: Extracting actionable
  knowledge for embodied agents.
\newblock In \emph{International Conference on Machine Learning}, pages
  9118--9147. PMLR, 2022.

\bibitem[Wang et~al.(2023{\natexlab{a}})Wang, Cai, Liu, Ma, and
  Liang]{wang2023describe}
Zihao Wang, Shaofei Cai, Anji Liu, Xiaojian Ma, and Yitao Liang.
\newblock Describe, explain, plan and select: Interactive planning with large
  language models enables open-world multi-task agents.
\newblock \emph{arXiv preprint arXiv:2302.01560}, 2023{\natexlab{a}}.

\bibitem[Mialon et~al.(2023)Mialon, Dess{\`\i}, Lomeli, Nalmpantis, Pasunuru,
  Raileanu, Rozi{\`e}re, Schick, Dwivedi-Yu, Celikyilmaz,
  et~al.]{mialon2023augmented}
Gr{\'e}goire Mialon, Roberto Dess{\`\i}, Maria Lomeli, Christoforos Nalmpantis,
  Ram Pasunuru, Roberta Raileanu, Baptiste Rozi{\`e}re, Timo Schick, Jane
  Dwivedi-Yu, Asli Celikyilmaz, et~al.
\newblock Augmented language models: a survey.
\newblock \emph{arXiv preprint arXiv:2302.07842}, 2023.

\bibitem[Qin et~al.(2023)Qin, Hu, Lin, Chen, Ding, Cui, Zeng, Huang, Xiao, Han,
  et~al.]{qin2023tool}
Yujia Qin, Shengding Hu, Yankai Lin, Weize Chen, Ning Ding, Ganqu Cui, Zheni
  Zeng, Yufei Huang, Chaojun Xiao, Chi Han, et~al.
\newblock Tool learning with foundation models.
\newblock \emph{arXiv preprint arXiv:2304.08354}, 2023.

\bibitem[Parisi et~al.(2022)Parisi, Zhao, and Fiedel]{parisi2022talm}
Aaron Parisi, Yao Zhao, and Noah Fiedel.
\newblock Talm: Tool augmented language models.
\newblock \emph{arXiv preprint arXiv:2205.12255}, 2022.

\bibitem[Schick et~al.(2023)Schick, Dwivedi-Yu, Dess{\`\i}, Raileanu, Lomeli,
  Zettlemoyer, Cancedda, and Scialom]{schick2023toolformer}
Timo Schick, Jane Dwivedi-Yu, Roberto Dess{\`\i}, Roberta Raileanu, Maria
  Lomeli, Luke Zettlemoyer, Nicola Cancedda, and Thomas Scialom.
\newblock Toolformer: Language models can teach themselves to use tools.
\newblock \emph{arXiv preprint arXiv:2302.04761}, 2023.

\bibitem[Lu et~al.(2023)Lu, Peng, Cheng, Galley, Chang, Wu, Zhu, and
  Gao]{lu2023chameleon}
Pan Lu, Baolin Peng, Hao Cheng, Michel Galley, Kai-Wei Chang, Ying~Nian Wu,
  Song-Chun Zhu, and Jianfeng Gao.
\newblock Chameleon: Plug-and-play compositional reasoning with large language
  models.
\newblock \emph{arXiv preprint arXiv:2304.09842}, 2023.

\bibitem[Shen et~al.(2023)Shen, Song, Tan, Li, Lu, and
  Zhuang]{shen2023hugginggpt}
Yongliang Shen, Kaitao Song, Xu~Tan, Dongsheng Li, Weiming Lu, and Yueting
  Zhuang.
\newblock Hugginggpt: Solving ai tasks with chatgpt and its friends in
  huggingface.
\newblock \emph{arXiv preprint arXiv:2303.17580}, 2023.

\bibitem[Wu et~al.(2023)Wu, Yin, Qi, Wang, Tang, and Duan]{wu2023visual}
Chenfei Wu, Shengming Yin, Weizhen Qi, Xiaodong Wang, Zecheng Tang, and Nan
  Duan.
\newblock Visual chatgpt: Talking, drawing and editing with visual foundation
  models.
\newblock \emph{arXiv preprint arXiv:2303.04671}, 2023.

\bibitem[Yang et~al.(2023)Yang, Song, Li, Zhao, Ge, Li, and
  Shan]{yang2023gpt4tools}
Rui Yang, Lin Song, Yanwei Li, Sijie Zhao, Yixiao Ge, Xiu Li, and Ying Shan.
\newblock Gpt4tools: Teaching large language model to use tools via
  self-instruction.
\newblock \emph{arXiv preprint arXiv:2305.18752}, 2023.

\bibitem[Yao et~al.(2022)Yao, Zhao, Yu, Du, Shafran, Narasimhan, and
  Cao]{yao2022react}
Shunyu Yao, Jeffrey Zhao, Dian Yu, Nan Du, Izhak Shafran, Karthik Narasimhan,
  and Yuan Cao.
\newblock React: Synergizing reasoning and acting in language models.
\newblock \emph{arXiv preprint arXiv:2210.03629}, 2022.

\bibitem[Li et~al.(2016)Li, Chou, Zhou, and Luo]{li2016rest}
Li~Li, Wu~Chou, Wei Zhou, and Min Luo.
\newblock Design patterns and extensibility of rest api for networking
  applications.
\newblock \emph{IEEE Transactions on Network and Service Management},
  13\penalty0 (1):\penalty0 154--167, 2016.

\bibitem[SmartBear(2023)]{swagger}
SmartBear.
\newblock Swagger, 2023.
\newblock URL \url{https://swagger.io/}.

\bibitem[Paranjape et~al.(2023)Paranjape, Lundberg, Singh, Hajishirzi,
  Zettlemoyer, and Ribeiro]{paranjape2023art}
Bhargavi Paranjape, Scott Lundberg, Sameer Singh, Hannaneh Hajishirzi, Luke
  Zettlemoyer, and Marco~Tulio Ribeiro.
\newblock Art: Automatic multi-step reasoning and tool-use for large language
  models.
\newblock \emph{arXiv preprint arXiv:2303.09014}, 2023.

\bibitem[Sur{\'\i}s et~al.(2023)Sur{\'\i}s, Menon, and
  Vondrick]{suris2023vipergpt}
D{\'\i}dac Sur{\'\i}s, Sachit Menon, and Carl Vondrick.
\newblock Vipergpt: Visual inference via python execution for reasoning.
\newblock \emph{arXiv preprint arXiv:2303.08128}, 2023.

\bibitem[Patil et~al.(2023)Patil, Zhang, Wang, and Gonzalez]{patil2023gorilla}
Shishir~G Patil, Tianjun Zhang, Xin Wang, and Joseph~E Gonzalez.
\newblock Gorilla: Large language model connected with massive apis.
\newblock \emph{arXiv preprint arXiv:2305.15334}, 2023.

\bibitem[Tang et~al.(2023)Tang, Deng, Lin, Han, Liang, and
  Sun]{tang2023toolalpaca}
Qiaoyu Tang, Ziliang Deng, Hongyu Lin, Xianpei Han, Qiao Liang, and Le~Sun.
\newblock Toolalpaca: Generalized tool learning for language models with 3000
  simulated cases.
\newblock \emph{arXiv preprint arXiv:2306.05301}, 2023.

\bibitem[Li et~al.(2023)Li, Song, Yu, Yu, Li, Huang, and Li]{li2023apibank}
Minghao Li, Feifan Song, Bowen Yu, Haiyang Yu, Zhoujun Li, Fei Huang, and
  Yongbin Li.
\newblock Api-bank: A benchmark for tool-augmented llms.
\newblock \emph{arXiv preprint arXiv:2304.08244}, 2023.

\bibitem[Wang et~al.(2023{\natexlab{b}})Wang, Li, Chen, Zhu, Lin, Cao, Liu,
  Liu, and Sui]{wang2023large}
Peiyi Wang, Lei Li, Liang Chen, Dawei Zhu, Binghuai Lin, Yunbo Cao, Qi~Liu,
  Tianyu Liu, and Zhifang Sui.
\newblock Large language models are not fair evaluators.
\newblock \emph{arXiv preprint arXiv:2305.17926}, 2023{\natexlab{b}}.

\bibitem[Masse(2011)]{masse2011rest}
Mark Masse.
\newblock \emph{REST API design rulebook: designing consistent RESTful web
  service interfaces}.
\newblock " O'Reilly Media, Inc.", 2011.

\bibitem[Shinn et~al.(2023)Shinn, Cassano, Labash, Gopinath, Narasimhan, and
  Yao]{shinn2023reflexion}
Noah Shinn, Federico Cassano, Beck Labash, Ashwin Gopinath, Karthik Narasimhan,
  and Shunyu Yao.
\newblock Reflexion: Language agents with verbal reinforcement learning.
\newblock \emph{arXiv preprint arXiv:2303.11366}, 2023.

\end{thebibliography}
}

\newpage
\appendix

\section{RESTful APIs and OAS}
RESTful API (Representational State Transfer API) is an architectural style used for designing networked applications.
It is based on the principles of Representational State Transfer (REST) and is widely used for building web services.\citep{masse2011rest,li2016rest}. 
In a RESTful API, resources (such as data objects or services) are represented as URLs (Uniform Resource Locators), known as endpoints. 
These endpoints are accessed over the HTTP protocol, and different HTTP methods (\texttt{GET}, \texttt{POST}, etc.) are used to perform operations on the resources.
There are millions of RESTful APIs available on Internet, such as Spotify, Twitter, Gmail, etc.

The OpenAPI Specification (OAS), formerly known as Swagger, is a specification for defining and documenting RESTful APIs  \citep{swagger}. 
It provides a standardized way to describe the structure, functionality, and behavior of an API, making it easier for developers to understand and interact with the API.
The OpenAPI Specification is written in JSON or YAML format and consists of a set of rules and conventions that define the endpoints, request/response formats, parameters, authentication methods, and other details of the API.
More specifically, an OAS consists of the following aspects for each API endpoint:
\begin{itemize}
    \item \textbf{API Path}: a relative path to an individual API endpoint, e.g., \texttt{/\{person\_id\}/details}.
    \item \textbf{API Description}: what the API does, how it works, and any potential errors or exceptions that may be raised.
    \item \textbf{Request Method}: the desired action to be performed for the API, e.g., \texttt{GET}, \texttt{POST}, \texttt{DELETE}.
    \item \textbf{Parameter List}: parameter name, parameter description, data type, default value, optional values of each parameter for the API.
    \item \textbf{Response Schema}: the schema of the response of the API. This information can assist the response parser to extract useful information from the JSON response.
    \item \textbf{Response Example (Optional)}: an example of a API call which can help demonstrate what the API will response.
    \item \textbf{Error and Exception}: potential error codes and their corresponding descriptions.
\end{itemize}

\begin{figure}[t]
    \centering
    \includegraphics[width=0.6\linewidth]{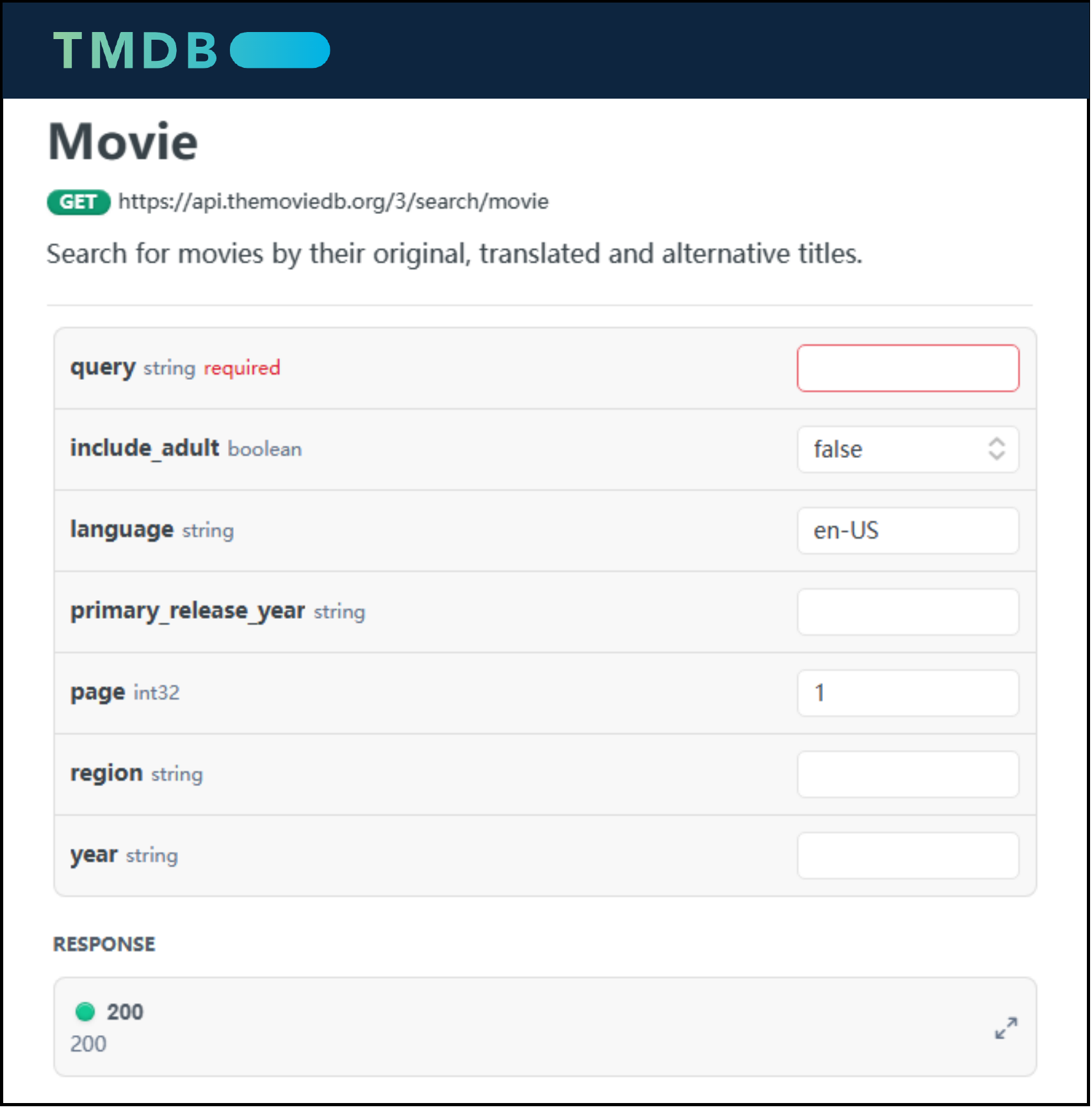}
    \caption{
        A RESTful API from TMDB.
    }
    \label{fig:rest_case}
\end{figure}

\begin{figure*}[t]
    \centering
    \includegraphics[width=1.0\linewidth]{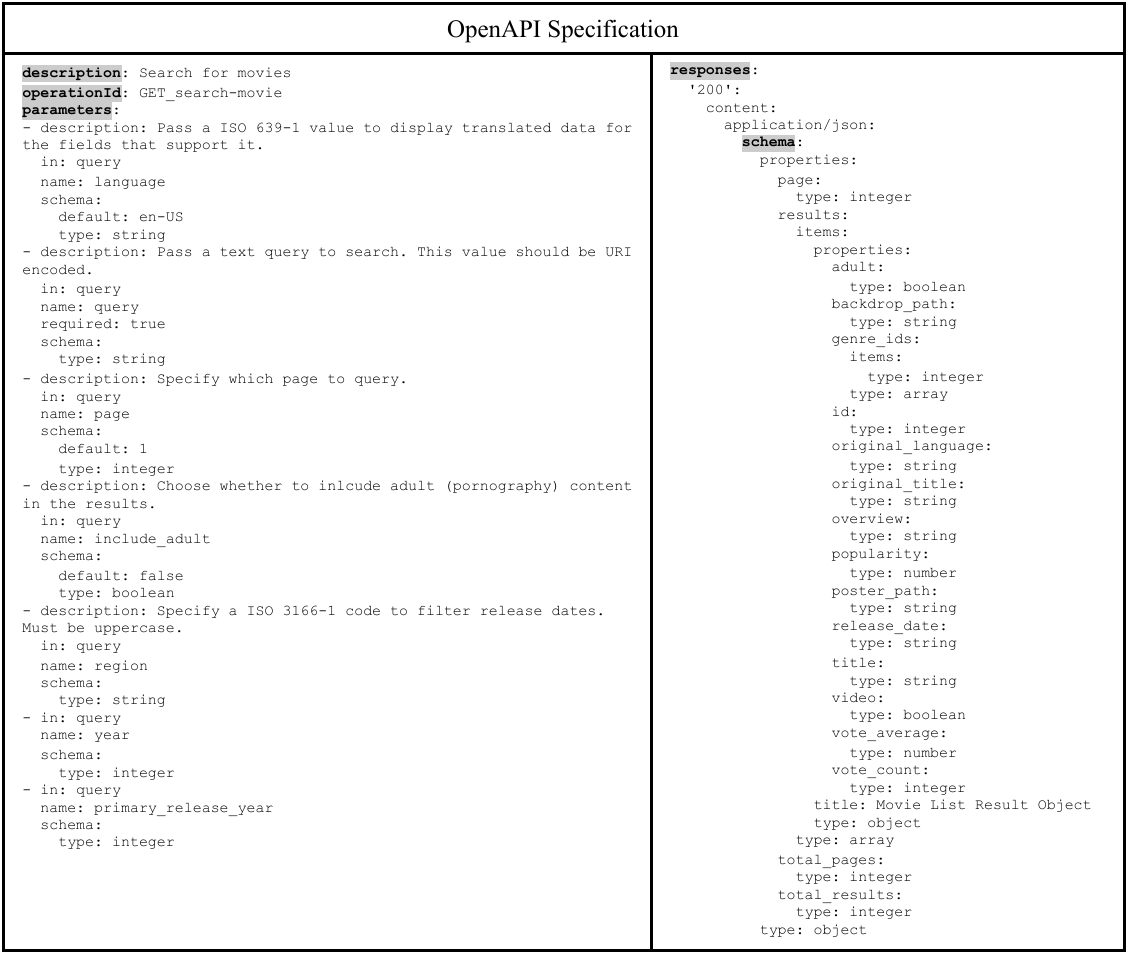}
    \caption{
        The OpenAPI Specification (OAS) of the API endpoint in Figure \ref{fig:rest_case}.
    }
    \label{fig:oas_case}
\end{figure*}

We provide an example of an OAS description of an API endpoint in Figure \ref{fig:rest_case} and Figure \ref{fig:oas_case}. 

\begin{figure*}[t!]
    \centering
    \includegraphics[width=1.0\linewidth]{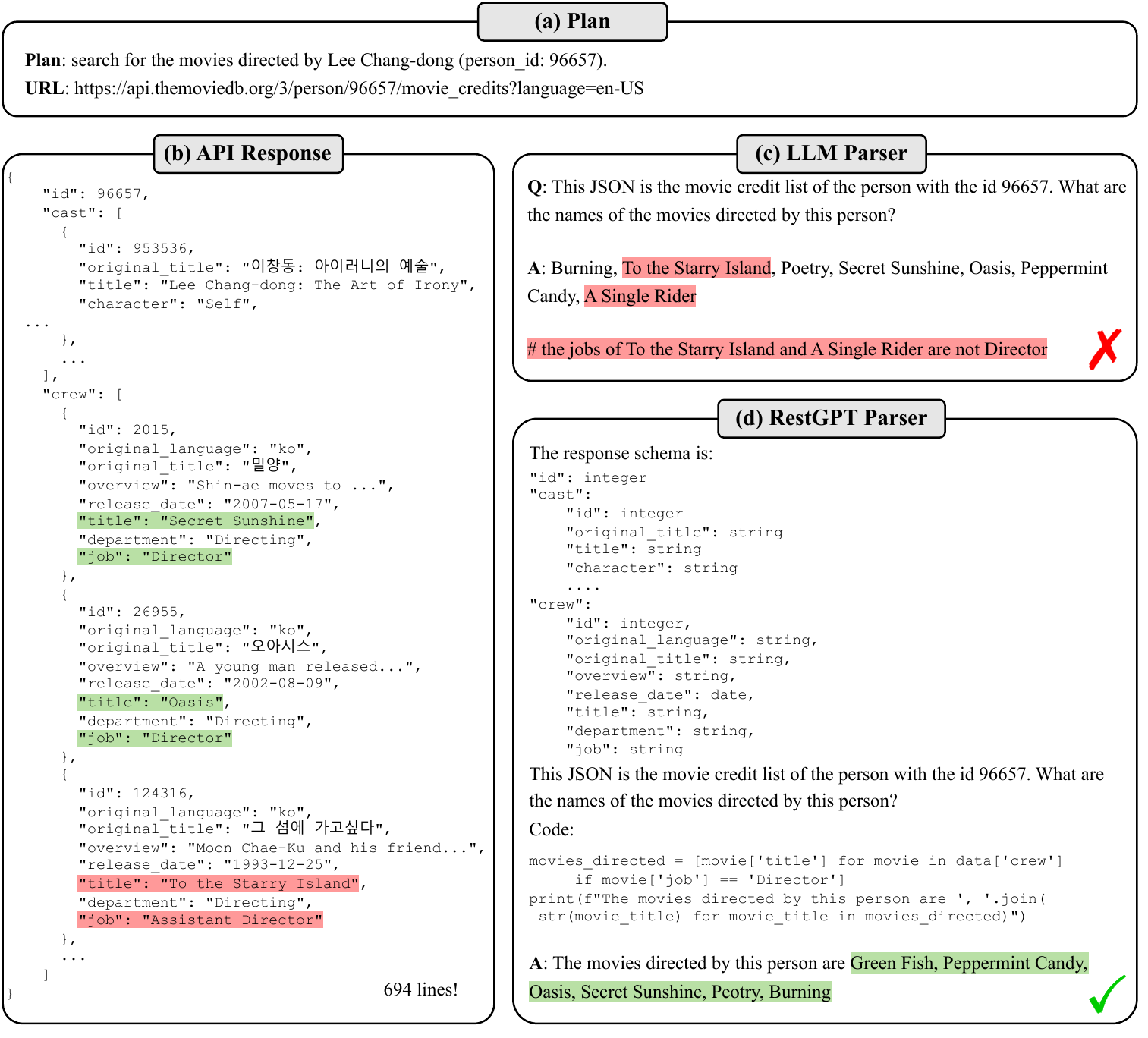}
    \caption{
        Case study on the response parser.
        The purpose of response parsing is to extract required information from the API response (b) according to the plan (a).
        We compare our proposed parser (d) with directly prompting an LLM as a parser (c).
    }
    \label{fig:parser_case}
\end{figure*}

\section{Baselines}
To demonstrate the effectiveness of \ourmethod{}, we compare it with four recent baselines:

\textbf{Offline introspective method} \citep{qin2023tool} is widely used in work on tool-augmented LLMs, such as HuggingGPT \citep{shen2023hugginggpt} and Chameleon \citep{lu2023chameleon}. 
This method directly generates multi-step plans for tool use and then sequentially executes the API plan. 
Due to the absence of API feedback, offline method cannot adjust the plan to adapt the unforeseen situations and errors in complex scenarios.

\textbf{DEPS} \citep{wang2023describe}, or ``Describe, Explain, Plan and Select'', is an interactive planning approach based on LLMs to improve the long-term planning ability on open-world problems.
DEPS will iteratively update the plan based on the environment feedback until the task is finished.

\textbf{ReAct} \citep{yao2022react} generates the chain-of-thought and actions in an interleaved manner.
The LLMs will reason about the current situation to make a better subsequent plan.

\textbf{Reflexion} \citep{shinn2023reflexion} is inspired from the ``trial and error'' of human learning process. 
It verbally reflect on task feedback signals, then maintain the reflective text in an episodic memory to induce better planning in subsequent trials.

Since some methods are not originally designed for tool/API usage, we reproduce them and add the API executor we proposed to make them able to call RESTful APIs.

\section{Case Study on the Executor}

Figure \ref{fig:parser_case} illustrates how \ourmethod{} parses JSON responses of RESTful APIs.
We compare \ourmethod{} parser with directly prompting an LLM as a parser.
As shown, responses from RESTful APIs are usually in lengthy and complex JSON format (694 lines in Figure \ref{fig:parser_case} (b))
The \ourmethod{} parser generates information extraction code based on the response schema in the OAS to avoid parsing lengthy JSON files directly, significantly improving the efficiency and accuracy of response parsing.

\section{More Cases on RestBench}

In this section, we show more cases of \ourmethod{} accomplishing complex instructions in two scenarios of RestBench.
Table \ref{tab:case1} and \ref{tab:case2} show two cases of the TMDB scenario.
Table \ref{tab:case3} and \ref{tab:case4} show two cases of the Spotify scenario.

\begin{table*}[t]
    \centering
    \fontsize{10.0pt}{\baselineskip}\selectfont
    \begin{boxedminipage}{1.0\columnwidth}
    
    \myboxx{\centering$\triangleright$ \textit{Instruction of TMDB}}
Give me the number of movies directed by Sofia Coppola.
\myboxx{\centering$\triangleright$ \textit{\ourmethod{}}}
\textbf{Planner:} search for the director ``Sofia Coppola''\\
\\
\textbf{API Selector:} GET /search/person to search for the director ``Sofia Coppola''\\
\\
\textbf{Caller:} \\
I need to make a GET request to the /search/person endpoint with the query parameter set to ``Sofia Coppola''.\\
Operation: GET\\
Input: \{\\
\ttab``url'': ``https://api.themoviedb.org/3/search/person'',\\
\ttab``params'': \{\\
\ttab\ttab``query'': ``Sofia Coppola''\\
\ttab\},\\
\ttab``description'': ``The API response is a list of people matching the query.'',\\
\ttab``output\_instructions'': ``What is the id of Sofia Coppola?''\\
\}\\
\\
\textbf{Parser:} The id of the person is 1769\\
\\
\textbf{Caller:} Successfully call GET /search/person to search for the director ``Sofia Coppola''. The id of Sofia Coppola is 1769.\\
\\
\textbf{Planner:} Get the number of movies directed by Sofia Coppola (1769)\\
\\
\textbf{API Selector:} GET /person/1769/movie\_credits to get the number of movies directed by Sofia Coppola (id 1769)\\
\\
\textbf{Caller:} \\
I need to call GET /person/1769/movie\_credits to get the number of movies directed by Sofia Coppola (id 1769).\\
Operation: GET\\
Input: \{\\
\ttab``url'': ``https://api.themoviedb.org/3/person/1769/movie\_credits'',\\
\ttab``description'': ``The API response is the movie credit list of Sofia Coppola (id 1769)'',\\
\ttab``output\_instructions'': ``What are the names and ids of the movies directed by this person?''\\
\}\\
\\
\textbf{Parser:} The movies directed by Sofia Coppola are Lost in Translation (153), The Virgin Suicides (1443), Marie Antoinette (1887), Somewhere (39210), Lick the Star (92657), The Bling Ring (96936), A Very Murray Christmas (364067), Bed, Bath and Beyond (384947), The Beguiled (399019), On the Rocks (575417), The Custom of the Country (758422), New York City Ballet (826970), Priscilla (1020006), Un matin partout dans le monde (835318)\\
\\
\textbf{Caller:} Successfully call GET /person/1769/movie\_credits to get the movie credit list of Sofia Coppola (id 1769). The movies directed by Sofia Coppola are Lost in Translation (153), The Virgin Suicides (1443), Marie Antoinette (1887), Somewhere (39210), Lick the Star (92657), The Bling Ring (96936), A Very Murray Christmas (364067), Bed, Bath and Beyond (384947), The Beguiled (399019), On the Rocks (575417), The Custom of the Country (758422), New York City Ballet (826970), Priscilla (1020006), Un matin partout dans le monde (835318).\\
\\
\textbf{Planner:} I am finished executing a plan and have the information the user asked for or the data the used asked to create. Final Answer: Sofia Coppola has directed 14 movies.
    \end{boxedminipage}
    \caption{Case 1 for TMDB.}
    \label{tab:case1}
\end{table*}

\begin{table*}[t]
    \centering
    \resizebox{1.0\linewidth}{!}{
    \fontsize{9.5pt}{\baselineskip}\selectfont
    \begin{boxedminipage}{1.0\columnwidth}
    
    \myboxx{\centering$\triangleright$ \textit{Instruction of TMDB}}
Who is the director of Leonardo DiCaprio's latest movie?
\myboxx{\centering$\triangleright$ \textit{\ourmethod{}}}
\textbf{Planner:} search for the latest movie of Leonardo DiCaprio\\
\textbf{API Selector:} GET /search/person to get the id of Leonardo DiCaprio\\
\textbf{Caller:} \\
I need to make a GET request to the /search/person endpoint with the query parameter set to ``Leonardo DiCaprio''\\
Operation: GET\\
Input: \{\\
\ttab``url'': ``https://api.themoviedb.org/3/search/person'',\\
\ttab``params'': \{``query'': ``Leonardo DiCaprio''\},\\
\ttab``description'': ``The API response is a list of people.'',\\
\ttab``output\_instructions'': ``What is the id of Leonardo DiCaprio?''\\
\}\\
\textbf{Parser:} The id of the person is 6193\\
\textbf{Caller:} Successfully call GET /search/person to search for the person ``Leonardo DiCaprio''. The id of Leonardo DiCaprio is 6193.\\
\\
\textbf{Planner:} Continue. search for the latest movie of Leonardo DiCaprio (6193)\\
\textbf{API Selector:} GET /person/6193/movie\_credits to get the latest movie of Leonardo DiCaprio (6193)\\
\textbf{Caller:} \\
I need to call GET /person/6193/movie\_credits to get the latest movie of Leonardo DiCaprio (6193).\\
Operation: GET\\
Input: \{\\
\ttab``url'': ``https://api.themoviedb.org/3/person/6193/movie\_credits'',\\
\ttab``description'': ``The API response is the movie credit list of Leonardo DiCaprio (id 6193)'',\\
\ttab``output\_instructions'': ``What is the name and id of the latest movie of Leonardo DiCaprio (6193)?''\\
\}\\
\textbf{Parser:} The name of the latest movie of Leonardo DiCaprio is Killers of the Flower Moon and the id is 466420\\
\textbf{Caller:} Successfully call GET /person/6193/movie\_credits to get the movie credit list of Leonardo DiCaprio (id 6193). The name of the latest movie of Leonardo DiCaprio is Killers of the Flower Moon and the id is 466420.\\
\\
\textbf{Planner:} search for the director of Killers of the Flower Moon (466420)\\
\textbf{Caller:} \\
I need to call GET /movie/\{movie\_id\}/credits to get the director of the movie Killers of the Flower Moon (id 466420).\\
Operation: GET\\
Input: \{\\
\ttab``url'': ``https://api.themoviedb.org/3/movie/466420/credits'',\\
\ttab``description'': ``The API response is the cast and crew for the movie Killers of the Flower Moon (id 466420).'',\\
\ttab``output\_instructions'': ``What is the name and id of the director of the movie Killers of the Flower Moon (id 466420)?''\\
\}\\
\textbf{Parser:} The name of the director of the movie Killers of the Flower Moon (id 466420) is Martin Scorsese (1032)\\
\textbf{Caller:} Successfully call GET /movie/466420/credits to get the director of the movie Killers of the Flower Moon (id 466420). The name and id of the director of the movie Killers of the Flower Moon (id 466420) is Martin Scorsese (1032).\\
\\
\textbf{Planner:} I am finished executing a plan and have the information the user asked for or the data the used asked to create. Martin Scorsese (1032) is the director of Leonardo DiCaprio's latest movie Killers of the Flower Moon (466420).
    \end{boxedminipage}
    }
    \caption{Case 2 for TMDB.}
    \label{tab:case2}
\end{table*}

\begin{table*}[t]
    \centering
    \resizebox{1.0\linewidth}{!}{
    \fontsize{10.0pt}{\baselineskip}\selectfont
    \begin{boxedminipage}{1.0\columnwidth}
    
    \myboxx{\centering$\triangleright$ \textit{Instruction of Spotify}}
Set the volume to 60 and skip to the next track.
\myboxx{\centering$\triangleright$ \textit{\ourmethod{}}}
\textbf{Planner:} set the volume to 60\\
\\
\textbf{API Selector:} PUT /me/player/volume to set the volume to 60\\
\\
\textbf{Caller:} \\
I need to make a PUT request to the /me/player/volume endpoint with the volume\_percent parameter set to 60.\\
Operation: PUT\\
Input: \{\\
\ttab``url'': ``https://api.spotify.com/v1/me/player/volume'',\\
\ttab``params'': \{\\
\ttab\ttab``volume\_percent'': ``60''\\
\ttab\},\\
\ttab``description'': ``Set the volume for the current playback device.''\\
\}\\
\\
\textbf{Parser:} The volume for the current playback device has been set to 60\%.\\
\\
\textbf{Caller:} Successfully call PUT /me/player/volume to set the volume to 60. The volume for the current playback device has been set to 60\%.\\
\\
\textbf{Planner:} skip to the next track\\
\\
\textbf{API Selector:} POST /me/player/next to skip to the next track\\
\\
\textbf{Caller:} \\
I need to make a POST request to the /me/player/next endpoint to skip to the next track.\\
Operation: POST\\
Input: \{\\
\ttab``url'': ``https://api.spotify.com/v1/me/player/next'',\\
\ttab``description'': ``Skips to next track in the user's queue.'',\\
\}\\
\\
\textbf{Parser:} The API call successfully skipped to the next track in the user's queue.\\
\\
\textbf{Caller:} Successfully call POST /me/player/next to skip to the next track.\\
\\
\textbf{Planner:} I am finished executing a plan and completed the user's instructions. I have set the volume to 60 and skipped to the next track.
    \end{boxedminipage}
    }
    \caption{Case 1 for Spotify.}
    \label{tab:case3}
\end{table*}

\begin{table*}[t]
    \centering
    \resizebox{1.0\linewidth}{!}{
    \fontsize{10.0pt}{\baselineskip}\selectfont
    \begin{boxedminipage}{1.0\columnwidth}
    
    \myboxx{\centering$\triangleright$ \textit{Instruction of Spotify}}
Add Summertime Sadness by Lana Del Rey in my first playlist.
\myboxx{\centering$\triangleright$ \textit{\ourmethod{}}}
\textbf{Planner:} Get the id of my first playlist\\
\textbf{API Selector:} GET /me/playlists to get the list of playlists owned or followed by the current Spotify user\\
\textbf{Caller:} \\
I need to use the GET method and generate the parameters according to the API documentation.\\
Operation: GET\\
Input: \{\\
\ttab``url'': ``https://api.spotify.com/v1/me/playlists'',\\
\ttab``description'': ``Get a list of the playlists owned or followed by the current Spotify user.'',\\
\ttab``params'': \{``limit'': ``1'', ``offset'': ``0''\},\\
\ttab``output\_instructions'': ``What are the ids and names of the playlists?''\\
\}\\
\textbf{Parser:} The id and name of the first playlist in the list is 6GHT and My Pop\\
\textbf{Caller:} Successfully call GET /me/playlists to get the list of playlists owned or followed by the current Spotify user. The id and name of the first playlist in the list is 6GHT and My Pop.\\
\\
\textbf{Planner:} Search for Summertime Sadness by Lana Del Rey\\
\textbf{API Selector:} GET /search to search for the track ``Summertime Sadness'' by Lana Del Rey\\
\textbf{Caller:} \\
I need to call GET /search with the appropriate parameters to search for the track ``Summertime Sadness'' by Lana Del Rey.\\
Operation: GET\\
Input: \{\\
\ttab``url'': ``https://api.spotify.com/v1/search'',\\
\ttab``params'': \{``q'': ``track:Summertime Sadness artist:Lana Del Rey'', ``type'': ``track'', ``limit'': ``1'', ``offset'': ``0''\},\\
\ttab``description'': ``The API response is a track object for the track 'Summertime Sadness' by Lana Del Rey.'',\\
\ttab``output\_instructions'': ``What is the id of the track?''\\
\}\\
\textbf{Parser:} The id of the track is 1Ist\\
\textbf{Caller:} Successfully call GET /search to search for the track ``Summertime Sadness'' by Lana Del Rey. The id of the track is 1Ist.\\
\\
\textbf{Planner:} Add Summertime Sadness by Lana Del Rey (1Ist) in my first playlist (6GHT)\\
\textbf{Caller:} \\
I need to make a POST request to add the track to the playlist.\\
Operation: POST\\
Input: \{\\
\ttab``url'': ``https://api.spotify.com/v1/playlists/6GHT/tracks'',\\
\ttab``data'': \{``uris'': [``spotify:track:1Ist'']\},\\
\ttab``description'': ``The API response is the track added to the playlist.'',\\
\ttab``output\_instructions'': ``What is the id of the snapshot of the playlist?''\\
\}\\
\textbf{Parser:} The snapshot id of the playlist is ``NCxj''.\\
\textbf{Caller:} Successfully call POST /playlists/6GHT/tracks to add the track ``Summertime Sadness'' (1Ist) to the playlist ``My Pop'' (6GHT).
\\
\textbf{Planner:} I am finished executing a plan and have the data the used asked to create. I have added Summertime Sadness by Lana Del Rey in your first playlist My Pop.
    \end{boxedminipage}
    }
    \caption{Case 2 for Spotify. Note that all IDs provided here are fake due to copyright restrictions.}
    \label{tab:case4}
\end{table*}

\section{Failure Cases}

We also present two failure cases of \ourmethod{}.
In the first case (Table \ref{tab:fail_case_1}), the API selector hallucinate to make up the in-path parameter \texttt{tv\_id} for the API.
The second failure case (Table \ref{tab:fail_case_2}) involved multiple issues.
Firstly, the executor correctly provided the desired query result, while the planner continued to generate a redundant plan.
Additionally, the response parser generated incorrect parsing code and got the name of the album instead of the track.
As a result, \ourmethod{} returned an incorrect answer.
These failure cases highlight the limitations of \ourmethod{} and the need for further improvement.

\begin{table*}[t]
    \centering
    \fontsize{10.0pt}{\baselineskip}\selectfont
    \begin{boxedminipage}{1.0\columnwidth}
    
    \myboxx{\centering$\triangleright$ \textit{Instruction of TMDB}}
Give me a photo belong to the second episode of the first season of the Witcher
\myboxx{\centering$\triangleright$ \textit{\ourmethod{}}}
\textbf{Planner:} search for the id of the second episode of the first season of the Witcher\\
\\
\textbf{API Selector:} \sethlcolor{error_red}\hl{GET /tv/63926/season/1/episode/2 to get the id of the second episode of the first season of the Witcher}\\
\\
\textbf{Caller:} \\
Operation: GET\\
Input: \{\\
\ttab``url'': ``https://api.themoviedb.org/3/tv/63926/season/1/episode/2'',\\
\ttab``description'': ``Get the id of the second episode of the first season of the Witcher'',\\
\ttab``output\_instructions'': ``What is the id of the second episode of the first season of the Witcher?''\\
\}\\
\\
\textbf{Parser:} The id of the second episode of the first season of the Witcher is 1132654\\
\\
\textbf{Caller:} Successfully call GET /tv/63926/season/1/episode/2 to get the id of the second episode of the first season of the Witcher. The id of the second episode of the first season of the Witcher is 1132654.\\
\\
\textbf{Planner:} search for the photo belong to the second episode of the first season of the Witcher (1132654)\\
\\
\textbf{...(Finally Failed)}
    \end{boxedminipage}
    \caption{A failure case for TMDB.}
    \label{tab:fail_case_1}
\end{table*}

\begin{table*}[t]
    \centering
    \fontsize{10.0pt}{\baselineskip}\selectfont
    \begin{boxedminipage}{1.0\columnwidth}
    
    \myboxx{\centering$\triangleright$ \textit{Instruction of Spotify}}
What is the name of the song I playing right now?
\myboxx{\centering$\triangleright$ \textit{\ourmethod{}}}
\textbf{Planner:} Get the current playing song\\
\\
\textbf{API Selector:} GET /me/player/currently-playing to get the track currently being played on the user's Spotify account\\
\\
\textbf{Caller:} \\
Operation: GET\\
Input: \{\\
\ttab``url'': ``https://api.spotify.com/v1/me/player/currently-playing'',\\
\ttab``params'': \{\\
\ttab\ttab``market'': ``US''\\
\ttab\},\\
\ttab``description'': ``Get the track currently being played on the user's Spotify account.'',\\
\ttab``output\_instructions'': ``What is the name and id of the track currently being played?''\\
\}\\
\\
\textbf{Parser:} The name of the track currently being played is LOVE~Destiny~, and its id is 5gAC\\
\\
\textbf{Caller:} Successfully call GET /me/player/currently-playing to get the track currently being played on the user's Spotify account. \sethlcolor{pass_green}\hl{The name of the track currently being played is LOVE~Destiny~, and its id is 5gAC.}\\
\\
\textbf{Planner:} \sethlcolor{error_red}\hl{Get the name of the song (5gAC)}\\
\\
\textbf{API Selector:} GET /tracks/5gAC to get the name of the song\\
\\
\textbf{Caller:} \\
I need to call GET /tracks/\{id\} to get the name of the song\\
Operation: GET\\
Input: \{\\
\ttab``url'': ``https://api.spotify.com/v1/tracks/5gAC'',\\
\ttab``description'': ``Get Spotify catalog information for the track LOVE~Destiny~'',\\
\ttab``output\_instructions'': ``What is the name of the song?''\\
\}\\
\\
\textbf{Parser:} \sethlcolor{error_red}\hl{The name of the song is A BEST}\\
\\
\textbf{Caller:} Successfully call GET /tracks/5gACkVtLdO8ITqbEF3ZfD1 to get the name of the song. The name of the song is A BEST.\\
\\
\textbf{Planner:} I am finished executing a plan and have the information the user asked for. \sethlcolor{error_red}\hl{The name of the song you are playing right now is A BEST.}
    \end{boxedminipage}
    \caption{A failure case for Spotify. Note that all IDs provided here are fake due to copyright restrictions.}
    \label{tab:fail_case_2}
\end{table*}

\section{Implementation Details}

In this section, we show the details of the prompt design in \ourmethod{}.
Some prompts are inspired by the OpenAPI agent implemented by LangChain\footnote{https://python.langchain.com/}.
The prompts of the planner, API selector, caller, response parser are shown in Table \ref{tab:planner_prompt}, \ref{tab:selector_prompt}, \ref{tab:caller_prompt}, \ref{tab:parser_prompt}, respectively.

\begin{table*}[t]
    \centering
    \fontsize{10.0pt}{\baselineskip}\selectfont
    \begin{boxedminipage}{1.0\columnwidth}
    
    \myboxx{\centering$\triangleright$ \textit{Prompt for the planner in \ourmethod{}}}
You are an agent that plans solution to user queries.\\
You should always give your plan in natural language.\\
Another model will receive your plan and find the right API calls and give you the result in natural language.\\
If you assess that the current plan has not been fulfilled, you can output "Continue" to let the API selector select another API to fulfill the plan.\\
If you think you have got the final answer or the user query has been fulfilled, just output the answer immediately. If the query has not been fulfilled, you should continue to output your plan.\\
The plan should be as specific as possible. It is better not to use pronouns in plan, but to use the corresponding results obtained previously. If you want to iteratively query something about items in a list, then the list and the elements in the list should also appear in your plan. The plan should be straightforward. If you want to search, sort or filter, you can put the condition in your plan.\\
\\
Starting below, you should follow this format:\\
\\
User query: the query a User wants help with related to the API.\\
Plan step 1: the first step of your plan for how to solve the query\\
API response: the result of executing the first step of your plan, including the specific API call made.\\
Plan step 2: based on the API response, the second step of your plan for how to solve the query. If the last step result is not what you want, you can output "Continue" to let the API selector select another API to fulfill the plan. For example, the last plan is "add a song (id xxx) in my playlist", but the last step API response is calling "GET /me/playlists" and getting the id of my playlist, then you should output "Continue" to let the API selector select another API to add the song to my playlist. Pay attention to the specific API called in the last step API response. If a inproper API is called, then the response may be wrong and you should give a new plan.\\
API response: the result of executing the second step of your plan\\
... (this Plan step n and API response can repeat N times)\\
Thought: I am finished executing a plan and have the information the user asked for or the data the used asked to create\\
Final Answer: the final output from executing the plan\\
\\
Example:\\
\texttt{\{in-context examples\}}\\
\\
Begin!\\
\\
User query: \texttt{\{query\}}\\
Plan step 1:
    \end{boxedminipage}
    \caption{The prompt of the planner.}
    \label{tab:planner_prompt}
\end{table*}

\begin{table*}[t]
    \centering
    \fontsize{10.0pt}{\baselineskip}\selectfont
    \begin{boxedminipage}{1.0\columnwidth}
    
    \myboxx{\centering$\triangleright$ \textit{Prompt for the API selector in \ourmethod{}}}
You are a planner that plans a sequence of RESTful API calls to assist with user queries against an API.\\
Another API caller will receive your plan call the corresponding APIs and finally give you the result in natural language.\\
The API caller also has filtering, sorting functions to post-process the response of APIs. Therefore, if you think the API response should be post-processed, just tell the API caller to do so.\\
If you think you have got the final answer, do not make other API calls and just output the answer immediately. For example, the query is search for a person, you should just return the id and name of the person.\\
\\
----\\
\\
Here are name and description of available APIs.\\
Do not use APIs that are not listed here.\\
\\
{endpoints}\\
\\
----\\
\\
Starting below, you should follow this format:\\
\\
Background: background information which you can use to execute the plan, e.g., the id of a person, the id of tracks by Faye Wong. In most cases, you must use the background information instead of requesting these information again. \\
User query: the query a User wants help with related to the API\\
API calling 1: the first api call you want to make. Note the API calling can contain conditions such as filtering, sorting, etc. If user query contains some filter condition, such as the latest, the most popular, the highest rated, then the API calling plan should also contain the filter condition. If you think there is no need to call an API, output "No API call needed." and then output the final answer according to the user query and background information.\\
API response: the response of API calling 1\\
Instruction: Another model will evaluate whether the user query has been fulfilled. If the instruction contains "continue", then you should make another API call following this instruction.\\
... (this API calling n and API response can repeat N times, but most queries can be solved in 1-2 step)\\
\\
Examples:\\
\\
\texttt{\{icl\_examples\}}\\
\\
Note, if the API path contains "\{\}", it means that it is a variable and you should replace it with the appropriate value. In most cases, the id value is in the background or the API response. Just copy the id faithfully. If the id is not in the background, instead of creating one, call other APIs to query the id.\\
\\
Begin!\\
\\
Background: \texttt{\{background\}}\\
User query: \texttt{\{plan\}}\\
API calling 1:
    \end{boxedminipage}
    \caption{The prompt of the API selector.}
    \label{tab:selector_prompt}
\end{table*}

\begin{table*}[t]
    \centering
    \fontsize{9.0pt}{\baselineskip}\selectfont
    \begin{boxedminipage}{1.0\columnwidth}
    
    \myboxx{\centering$\triangleright$ \textit{Prompt for the caller in \ourmethod{}}}
You are an agent that gets a sequence of API calls and given their documentation, should execute them and return the final response.\\
If you cannot complete them and run into issues, you should explain the issue. If you're able to resolve an API call, you can retry the API call. When interacting with API objects, you should extract ids for inputs to other API calls but ids and names for outputs returned to the User.\\
Your task is to complete the corresponding api calls according to the plan.\\
\\
Here is documentation of the API:\\
Base url: \texttt{\{api\_url\}}\\
Endpoints:\\
\texttt{\{api\_docs\}}\\
\\
If the API path contains "\{\}", it means that it is a variable and you should replace it with the appropriate value. For example, if the path is "/users/\{user\_id\}/tweets", you should replace "\{user\_id\}" with the user id. "\{" and "\}" cannot appear in the url.\\
\\
You can use http request method, i.e., GET, POST, DELETE, PATCH, PUT, and generate the corresponding parameters according to the API documentation and the plan.\\
The input should be a JSON string which has 3 base keys: url, description, output\_instructions\\
The value of "url" should be a string.\\
The value of "description" should describe what the API response is about. The description should be specific.\\
The value of "output\_instructions" should be instructions on what information to extract from the response, for example the id(s) for a resource(s) that the POST request creates. Note "output\_instructions" must be natural language and as verbose as possible! It cannot be "return the full response". Output instructions should faithfully contain the contents of the api calling plan and be as specific as possible. The output instructions can also contain conditions such as filtering, sorting, etc.\\
If you are using GET method, add "params" key, and the value of "params" should be a dict of key-value pairs.\\
If you are using POST, PATCH or PUT methods, add "data" key, and the value of "data" should be a dict of key-value pairs.\\
\\
Examples: \texttt{\{icl\_examples\}}\\
\\
I will give you the background information and the plan you should execute.\\
You should execute the plan faithfully and give the Final Answer as soon as you successfully call the planned APIs, don't get clever and make up steps that don't exist in the plan. Do not make up APIs that don't exist in the plan.\\
\\
Starting below, you must follow this format:\\
\\
Background: background information which you can use to execute the plan, e.g., the id of a person.\\
Plan: the plan of API calls to execute\\
Thought: you should always think about what to do\\
Operation: the request method to take, should be one of the following: GET, POST, DELETE, PATCH, PUT\\
Input: the input to the operation\\
Response: the output of the operation\\
Thought: I am finished executing the plan\\
Execution Result: based on the API response, the execution result of the API calling plan.\\
\\
Begin!\\
\\
Background: \texttt{\{background\}}\\
Plan: \texttt{\{api\_plan\}}\\
Thought:
    \end{boxedminipage}
    \caption{The prompt of the caller.}
    \label{tab:caller_prompt}
\end{table*}

\begin{table*}[t]
    \centering
    \fontsize{10.0pt}{\baselineskip}\selectfont
    \begin{boxedminipage}{1.0\columnwidth}
    
    \myboxx{\centering$\triangleright$ \textit{Prompt for the parser in \ourmethod{}}}
Here is an API response schema from an OAS and a query. \\
The API's response will follow the schema and be a JSON. \\
Assume you are given a JSON response which is stored in a python dict variable called 'data', your task is to generate Python code to extract information I need from the API response.\\
Note: I will give you 'data', do not make up one, just reference it in your code.\\
Please print the final result as brief as possible. If the result is a list, just print it in one sentence. Do not print each item in a new line.\\
Note you should generate only Python code.\\
DO NOT use fields that are not in the response schema.\\
\\
API: \texttt{\{api\_path\}}\\
API description: \texttt{\{api\_description\}}\\
Parameters or body for this API call:\\
\texttt{\{api\_param\}}\\
\\
Response JSON schema defined in the OAS:\\
\texttt{\{response\_schema\}}\\
\\
The response is about: \texttt{\{response\_description\}}\\
\\
Query: \texttt{\{query\}}\\
\\
The code you generate should satisfy the following requirements:\\
1. The code you generate should contain the filter in the query.\\
2. If the response is something about X, then the filter condition cannot include searching for X.\\
3. Do not use f-string in the print function. Use "format" instead.\\
4. Please print the final result as brief as possible. If the result is a list, just print it in one sentence. Do not print each item in a new line.\\
\\
Begin!\\
Python Code:
    \end{boxedminipage}
    \caption{The prompt of the parser.}
    \label{tab:parser_prompt}
\end{table*}

\end{document}